\documentclass[conference]{IEEEtran}
\IEEEoverridecommandlockouts
\usepackage{cite}
\usepackage{amsmath,amssymb,amsfonts}
\usepackage{algorithmic}
\usepackage{graphicx}
\usepackage{textcomp}
\usepackage[dvipsnames]{xcolor}
\usepackage{colortbl}
\usepackage{booktabs}
\usepackage{multirow}
\usepackage{multicol}
\usepackage{fancyhdr}
\usepackage{tikz}
\def\BibTeX{{\rm B\kern-.05em{\sc i\kern-.025em b}\kern-.08em
    T\kern-.1667em\lower.7ex\hbox{E}\kern-.125emX}}

\newcommand\copyrighttext{%
    This work has been submitted to the IEEE for possible publication. Copyright may be transferred without notice, after which this version may no longer be accessible.}
\newcommand\copyrightnotice{%
\begin{tikzpicture}[remember picture,overlay]
\node[anchor=south,yshift=65pt] at (current page.south) {\fbox{\parbox{\dimexpr\textwidth-\fboxsep-\fboxrule\relax}{\copyrighttext}}};
\end{tikzpicture}%
}

\begin{document}

\copyrightnotice

\title{Generating Valid and Natural Adversarial Examples with Large Language Models
}

\author{\IEEEauthorblockN{Zimu Wang$^{1,3}$, Wei Wang$^1$\IEEEauthorrefmark{1}\thanks{\IEEEauthorrefmark{1}Corresponding author.}, Qi Chen$^2$, Qiufeng Wang$^1$, Anh Nguyen$^3$}
\IEEEauthorblockA{$^1$\textit{School of Advanced Technology};$\quad$$^2$\textit{School of AI and Advanced Computing,} \\
\textit{Xi'an Jiaotong-Liverpool University, Suzhou, China} \\
$^3$\textit{Department of Computer Science, University of Liverpool, Liverpool, United Kingdom} \\
}
Zimu.Wang19@student.xjtlu.edu.cn, \{Wei.Wang03, Qi.Chen02, Qiufeng.Wang\}@xjtlu.edu.cn \\
Anh.Nguyen@liverpool.ac.uk
}

\maketitle

\begin{abstract}
Deep learning-based natural language processing (NLP) models, particularly pre-trained language models (PLMs), have been revealed to be vulnerable to adversarial attacks. However, the adversarial examples generated by many mainstream word-level adversarial attack models are neither valid nor natural, leading to the loss of semantic maintenance, grammaticality, and human imperceptibility. Based on the exceptional capacity of language understanding and generation of large language models (LLMs), we propose LLM-Attack, which aims at generating both valid and natural adversarial examples with LLMs. The method consists of two stages: word importance ranking (which searches for the most vulnerable words) and word synonym replacement (which substitutes them with their synonyms obtained from LLMs). Experimental results on the Movie Review (MR), IMDB, and Yelp Review Polarity datasets against the baseline adversarial attack models illustrate the effectiveness of LLM-Attack, and it outperforms the baselines in human and GPT-4 evaluation by a significant margin. The model can generate adversarial examples that are typically valid and natural, with the preservation of semantic meaning, grammaticality, and human imperceptibility. 
\end{abstract}

\begin{IEEEkeywords}
Adversarial attack, adversarial examples, text classification, large language models, natural language processing.
\end{IEEEkeywords}

\section{Introduction}

Deep learning-based natural language processing (NLP) models, particularly pre-trained language models (PLMs), have achieved the state-of-the-art performance in an extensive list of downstream tasks, such as sentiment analysis \cite{Wang-and-Gan-2023-Adversarial}, information extraction \cite{Wang-etal-2022-MAVEN-ERE}, and text summarisation \cite{Lu-etal-2023-Summarization}. Despite the success, numerous studies have revealed that PLMs are vulnerable to adversarial attacks, i.e. minor perturbations imposed on the original input may lead the models to make incorrect prediction \cite{Wang-and-Gan-2023-Adversarial,Chen-etal-2020-Adversarial}. Such phenomena facilitate the evaluation and improvement of robustness, security, and explainability of NLP models \cite{Chen-etal-2022-Adversarial}.

\begin{figure}[t!]
    \centering
    \includegraphics[width=0.9\linewidth]{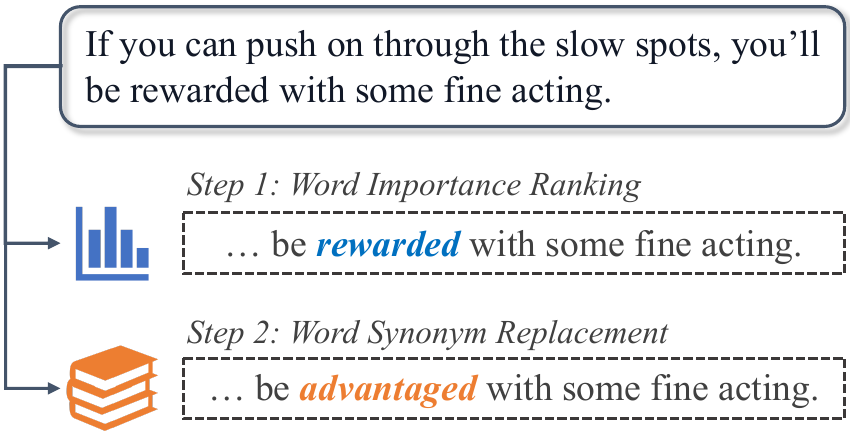}
    \caption{Overall procedure of generating word-level adversarial examples, consisting of two stages: 1) word importance ranking and 2) word synonym replacement.}
    \label{fig:previous-work}
\end{figure}

Existing research on textual adversarial attacks can be categorised onto four levels: character-level, word-level, sentence-level, and multi-level adversarial attacks. The word-level adversarial attack is one of the most popular paradigms, consisting of two stages: ranking tokens importance and replacing them with heuristic rules \cite{Fang-etal-2023-Decision-Making}, as shown in Fig. \ref{fig:previous-work}. As suggested in recent studies \cite{Jin-etal-2020-TextFooler,Dyrmishi-etal-2023-Valid-Natural}, effective adversarial examples should be \textbf{\textit{valid}} and \textbf{\textit{natural}}, with preservation of the following properties: 1) \textit{Human prediction consistency}: maintaining human perception while changing the models’ prediction; 2) \textit{Semantic similarity}: preserving semantic meaning of the original input; and 3) \textit{Language fluency}: ensuring grammatical correctness of the generated adversarial examples.

\begin{figure*}[t!]
    \centering
    \begin{tabular}{m{0.4cm}<{\centering}|m{16.8cm}<{}}\toprule
        {\small \textbf{\#}} & {\small \textbf{Original \& Adversarial Examples}} \\
        \midrule
        \multirow{2}{*}{\small 1} & {\small Rita Hayworth is just stunning at {\color{NavyBlue} \textbf{times}} and, for me, the only reason to watch this {\color{NavyBlue} \textbf{silly}} film.} \\
        & {\small Rita Hayworth is just stunning at {\color{gray} \textbf{length}} and, for me, the only reason to watch this {\color{BrickRed} \textbf{wonderful}} film.} \\
        \midrule
        \multirow{2}{*}{\small 2} & {\small Seeing the title of this movie “Stupid Teenagers {\color{NavyBlue} \textbf{Must Die}}” {\color{NavyBlue} \textbf{made}} me believe this was a spoof of some {\color{NavyBlue} \textbf{kind}}.} \\
        & {\small Seeing the title of this movie “Stupid Teenagers {\color{violet} \textbf{I Died}}” {\color{PineGreen} \textbf{had}} me believe this was a spoof of some {\color{gray} \textbf{work}}.} \\
        \midrule
        \multirow{2}{*}{\small 3} & {\small {\color{NavyBlue} \textbf{I love}} sci-fi and am {\color{NavyBlue} \textbf{willing}} to put up with a lot. Sci-fi movies/TV are usually {\color{NavyBlue} \textbf{underfunded}}, under-appreciated and misunderstood.} \\
        & {\small {\color{BurntOrange} \textbf{Me enjoys}} sci-fi and am {\color{BrickRed} \textbf{hard}} to put up with a lot. Sci-fi movies/TV are usually {\color{gray} \textbf{ridiculous}}, under-appreciated and misunderstood.} \\
        \bottomrule
    \end{tabular}
    \caption{Adversarial examples generated by the BAE adversarial attack model on the IMDB dataset. ({\color{NavyBlue} \textbf{Blue}}: selected words to replace, {\color{PineGreen} \textbf{Green}}: valid word replacements, {\color{BrickRed} \textbf{Red}}: antonyms, {\color{violet} \textbf{Violet}}: split named entities, {\color{BurntOrange} \textbf{Orange}}: grammatical errors, {\color{gray} \textbf{Gray}}: other unrelated replacements.)}
    \label{fig:BAE}
\end{figure*}

However, recent studies have revealed that many word-level adversarial examples do not meet the aforementioned requirements \cite{Dyrmishi-etal-2023-Valid-Natural,Chiang-and-Lee-2023-Synonym}, e.g. semantic meaning cannot be preserved, and grammatical errors are made during word substitution using WordNet \cite{Miller-1995-WordNet} and BERT \cite{Devlin-etal-2019-BERT}. As shown in Fig. \ref{fig:BAE}, the current word-level adversarial attack research faces the following unique challenges.

\begin{itemize}
    \item  \textbf{Semantic change} -- With only a small fraction of true synonyms in the selected candidate list, a word has a high likelihood of being altered to its antonyms or unrelated words \cite{Chiang-and-Lee-2023-Synonym}. 
    \item  \textbf{Named entity split} -- Many approaches rely heavily on word importance during the substitution process, where the change of words in named entities can dramatically change the original text meanings and make the adversarial examples perceptible. 
    \item  \textbf{Grammatical errors} -- Synonym-based word substitution often ignores the word senses, tenses, and possessive cases, while substitution of pronouns tends to replace them with other pronouns or their possessive forms, which could result in grammatical errors and human perceptibility.
\end{itemize}

Large language models (LLMs), such as ChatGPT\footnote{https://chat.openai.com/} and ChatGLM \cite{Zeng-etal-2023-GLM}, have demonstrated stunning capabilities in semantic understanding, text generation, and grammatical error correction \cite{Peng-etal-2023-ICL,Wu-etal-2023-Grammarly}. Such capabilities would assist in the generation of effective word-level adversarial examples. To address the problems mentioned above, we propose \textsc{LLM-Attack}, aiming at generating both valid and natural adversarial examples with LLMs. Following the mainstream paradigm of word-level adversarial attack, as shown in Fig. \ref{fig:previous-work}, the overall process of \textsc{LLM-Attack} consists of two stages: word importance ranking which searches for the most vulnerable words, and word synonym replacement which substitutes them with synonyms obtained from LLMs. With the exceptional capacity of language understanding and generation of LLMs, the synonym lists obtained are typically valid and natural, with the preservation of both semantic meaning and grammatical correctness. Experimental results on the Movie Review (MR) \cite{Pang-and-Lee-2005-MR}, IMDB \cite{Maas-etal-2011-IMDB}, and Yelp Review Polarity \cite{Zhang-etal-2015-AG-Yelp} datasets demonstrate the effectiveness of \textsc{LLM-Attack} against the baselines such as \textsc{TextFooler} \cite{Jin-etal-2020-TextFooler} and \verb|BAE| \cite{Garg-and-Ramakrishnan-2020-BAE}. It also outperforms the baselines in human and GPT-4 \cite{OpenAI-2023-GPT4} evaluation by a significant margin, indicating a higher level of effectiveness, validity, and naturalness of the generated adversarial examples.

The contributions of our work are summarised as follows:
\begin{itemize}
    \item We propose \textsc{LLM-Attack}, which aims to generate both valid and natural adversarial examples with LLMs. To the best of our knowledge, we are the first to incorporate LLMs in textual adversarial attack research.
    \item We evaluate the generated adversarial examples under three settings: automatic, human, and GPT-4 evaluations. To the best of our knowledge, this is the first work that evaluates the quality of adversarial examples using LLMs.
    \item Experimental results under automatic evaluation illustrate the effectiveness of \textsc{LLM-Attack}; it also outperforms the baselines under human and GPT-4 evaluation results by a significant margin.
\end{itemize}

\section{Related Work}
Adversarial attacks have been extensively studied over the past years with success in computer vision and speech domains, considering that deep neural networks (DNNs) are vulnerable to adversarial attacks. However, due to the discrete nature of natural languages, it is still far from ideal to perform successful adversarial attacks on NLP models \cite{Fang-etal-2023-Decision-Making}. 

Previous research on textual adversarial attacks can be divided onto four levels: character-level, word-level, sentence-level, and multi-level adversarial attacks. Some off-the-shelf toolkits have also been developed to generate adversarial examples against real-world models \cite{Morris-etal-2020-TextAttack,Zeng-etal-2021-OpenAttack}. Different from the character-level adversarial attack that manipulates the characters \cite{Gao-etal-2018-DeepWordBug,Eger-etal-2019-VIPER} and the sentence-level adversarial attack that creates new sentences with generative adversarial network (GAN) \cite{Zhao-etal-2018-GAN-Adversarial}, semantically equivalent rules \cite{Ribeiro-etal-2018-Rules}, and paraphrase networks \cite{Iyyer-etal-2018-Paraphrase-Adversarial}, word-level adversarial attack has been the most popular technique due to its potency in preserving the semantic similarity and imperceptibility of the adversarial examples. 

It usually consists of two stages: ranking importance of tokens and replacing them with heuristic rules \cite{Fang-etal-2023-Decision-Making}. In the first stage, the models iteratively search for the most vulnerable words to identify the modification positions, e.g. saliency-based ranking \cite{Fang-etal-2023-Decision-Making,Jin-etal-2020-TextFooler,Ren-etal-2019-PWWS} and gradient-based descent algorithms \cite{Li-etal-2019-TextBugger,Yoo-and-Qi-2021-A2T}; then the selected words are substituted with some pre-defined strategies, e.g. synonym vocabularies \cite{Ren-etal-2019-PWWS}, word embeddings \cite{Fang-etal-2023-Decision-Making,Jin-etal-2020-TextFooler}, and language models \cite{Garg-and-Ramakrishnan-2020-BAE,Li-etal-2020-BERTAttack}. By substituting synonyms for the most important words and imposing constraints during the replacements, such as maximum modification rate, part-of-speech (POS) consistency, stop words preservation, and word embedding distance, the generated adversarial examples have shown to downgrade the performance of well-trained text classifiers successfully.

However, recent investigation has revealed that current word-based adversarial examples are neither valid nor natural, with the possible loss of semantic maintenance, grammaticality, and human imperceptibility \cite{Dyrmishi-etal-2023-Valid-Natural,Chiang-and-Lee-2023-Synonym}. Different from the previous works, we leverage LLMs with exceptional capacities for language understanding and generation to generate both valid and natural adversarial examples.

\section{Methodology}

\begin{figure*}[t!]
    \centering
    \includegraphics[width=1\linewidth]{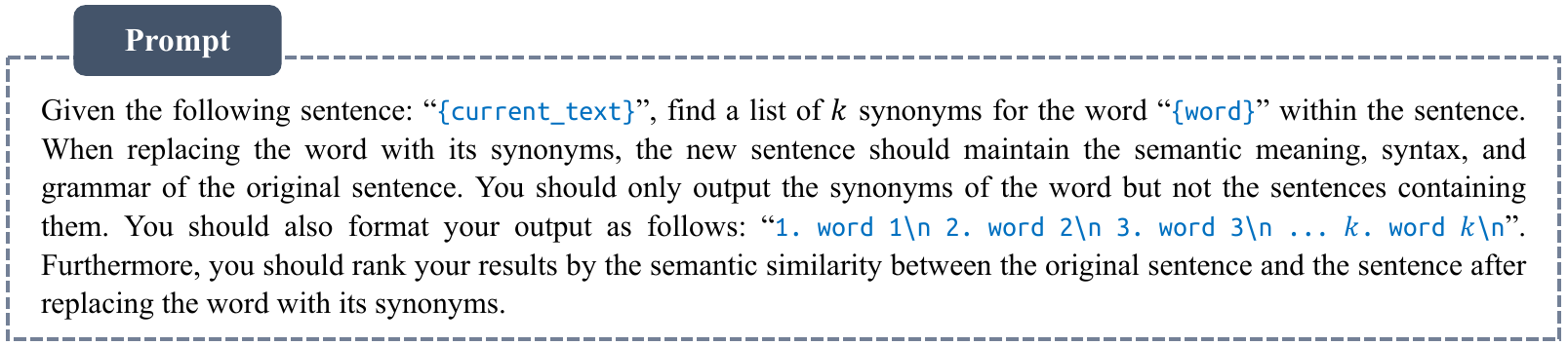}
    \caption{Prompt for obtaining the candidate synonyms of a given word using LLMs.}
    \label{fig:prompt}
\end{figure*}

\textsc{LLM-Attack} leverages LLMs to generate both valid and natural word-level adversarial examples and attempts to fool the trained NLP models. We focus on the text classification task and formulate it as follows: given a trained natural language classifier $\mathcal{F}: \mathcal{X} \rightarrow \mathcal{Y}$ that aims to correctly classify the input text $\mathbf{x}$ to the corresponding label $y_{true}$ with the maximum posterior probability:
\begin{equation}
    \text{arg max}_{y_i \in \mathcal{Y}} P(y_i|\mathbf{x}) = y_{\text{true}},
\end{equation}
we attack the classifier by introducing imperceptible perturbations, defined as $\mathbf{\Delta x}$, to craft an adversarial example $\mathbf{x}'$, for which $\mathcal{F}$ is expected to predict an incorrect label:
\begin{equation}
    \text{arg max}_{y_i \in \mathcal{Y}} P(y_i|\mathbf{x}') \neq y_{\text{true}}.
\end{equation}
We define the adversarial example in Eq. (\ref{eq:adv}):
\begin{equation}\label{eq:adv}
\begin{split}
    \mathbf{x}' = \mathbf{x} + \Delta \mathbf{x}, \quad ||\Delta \mathbf{x}||_p < \epsilon, \quad \mathbf{x}' \in \mathcal{X}, \\    
    \text{arg max}_{y_i \in \mathcal{Y}} P(y_i|\mathbf{x}) \neq \text{arg max}_{y_i \in \mathcal{Y}} P(y_i|\mathbf{x}'),
\end{split}
\end{equation}
in which $||\Delta \mathbf{x}||_p$ denotes the $p$-norm distance metric ($p = 0, 2,$ $\infty$):
\begin{equation}
    ||\Delta \mathbf{x}||_p = \sqrt[p]{\sum_{i=1}^n |x_i'-x_i|^p}.
\end{equation}

Following the procedure in Fig. \ref{fig:previous-work}, we explain the  two stages: (1) finding the most important words within the input sentence and (2) replacing the words with their synonyms by prompting LLMs in Secs. \ref{sec:wir} and \ref{sec:synonym}, respectively.

\subsection{Word Importance Ranking}\label{sec:wir}

We first rank the importance of words within the input sentence to find the most vulnerable ones and determine the word replacement order. Under the black-box scenario, only the logits output by the target models, such as fine-tuned PLMs, can be regarded as the supervision signal. To balance effectiveness and efficiency of word ranking, we adopt the masked language modeling (MLM) strategy to calculate the word importance. Specifically, let $\mathbf{x}$ denote the input sentence and $\mathbf{x}_i'$ denote the sentence after replacing the $i$-th word $w_i$ to \textsc{[UNK]}, as illustrated in Eqs. (\ref{eq:s}) and (\ref{eq:ss}):
\begin{align}
    \mathbf{x} &= [w_1, \cdots, w_{i-1}, w_i, w_{i+1}, \cdots, w_n],\label{eq:s}\\
    \mathbf{x_i'} &= [w_1, \cdots, w_{i-1}, \textsc{[UNK]}, w_{i+1}, \cdots, w_n],\label{eq:ss}
\end{align}
the word importance score for $w_i$, denoted as $I_{w_i}$, is defined as:
\begin{equation}
    I_{w_i} = P(y_{true}|\mathbf{x}) - P(y_{true}|\mathbf{x}_i').
\end{equation}

After obtaining the word importance score for all the words, we rank them according to $I_{w_i}$ in the descending order. We only take $\theta$ of the most important words for semantic preservation to create a candidate word list $L$.

\subsection{Word Synonym Replacement}\label{sec:synonym}

After determining the vulnerable words to be perturbed, we iteratively obtain the synonyms of the  words with LLMs using the prompt illustrated in Fig. \ref{fig:prompt}. This process aims at searching for $k$ synonyms of the vulnerable words that can potentially preserve the semantic meaning, syntax, and grammar of the original sentence, and ranking the output based on the semantic similarity between the original sentence and the one after perturbation. Following the previous work \cite{Fang-etal-2023-Decision-Making}, we also apply the constraints on repeat modification limitation, maximum modification rate, POS consistency, stop word preservation, and word embedding distance. We also employ the following additional constraints according to the research challenges in textual adversarial attacks.

\begin{enumerate}
    \item \textit{Named Entity and Pronoun Preservation}: As shown in Fig. \ref{fig:previous-work}, named entities (such as people, locations, and companies) are usually disregarded in prior adversarial attacks, and the alteration of named entities likely results in semantic changes, grammatical errors, and human perceptibility. For this reason, we employ spaCy\footnote{https://spacy.io/} to perform named entity recognition (NER) and skip the words that are part of the named entities during the word substitution procedure. Additionally, considering the fact that substitution of pronouns tends to cause grammatical errors, as shown in Fig. \ref{fig:previous-work}, we incorporate the pronouns to the stop word list to make sure they cannot be changed.
    \item \textit{Universal Sentence Encoder}: Although previous work concludes that the universal sentence encoder (USE) \cite{Cet-etal-2018-USE} is insensitive to invalid word substitutions, such as the substitution to mismatched synonyms and antonyms \cite{Chiang-and-Lee-2023-Synonym}, we employ USE to prevent LLMs from generating invalid outputs, such as redundant information apart from a list of words.
\end{enumerate}

\section{Experiments}

\subsection{Datasets and Evaluation Metrics}

Following the previous works in \cite{Jin-etal-2020-TextFooler,Garg-and-Ramakrishnan-2020-BAE}, we evaluated the performance of \textsc{LLM-Attack} on the text classification task. The datasets used for evaluation include both sentence-level (MR \cite{Pang-and-Lee-2005-MR}) and document-level (IMDB \cite{Maas-etal-2011-IMDB} and Yelp Review Polarity \cite{Zhang-etal-2015-AG-Yelp}) sentiment analysis datasets. The overall statistics of the datasets are shown in Table \ref{tab:datasets}. We randomly selected $500$ samples from the test set of each dataset, and reported both accuracy and attack success rate for each adversarial attack model.

\subsection{Baselines}

We compared \textsc{LLM-Attack} with two state-of-the-art adversarial attack models: 1) \textsc{TextFooler} \cite{Jin-etal-2020-TextFooler} is a simple but effective baseline that identifies the keywords by calculating the prediction change after iteratively deleting the words and selects the synonyms using the counter-fitted word embeddings. 2) \verb|BAE| \cite{Garg-and-Ramakrishnan-2020-BAE} determines the replaced words with the BERT-MLM approach after finding the vulnerable words with the deletion strategy. We selected the replacement strategy of \verb|BAE| due to its high sentiment accuracy and naturalness compared with the rest of the strategies. To guarantee fairness in the comparison study, we reported the baseline results under two settings: (1) the initial setting proposed by the authors, and (2) the setting proposed in this work, i.e. by applying the same constraints with \textsc{LLM-Attack}.

\subsection{Experimental Setup}

\begin{figure}[t!]
    \centering
    \includegraphics[width=1\linewidth]{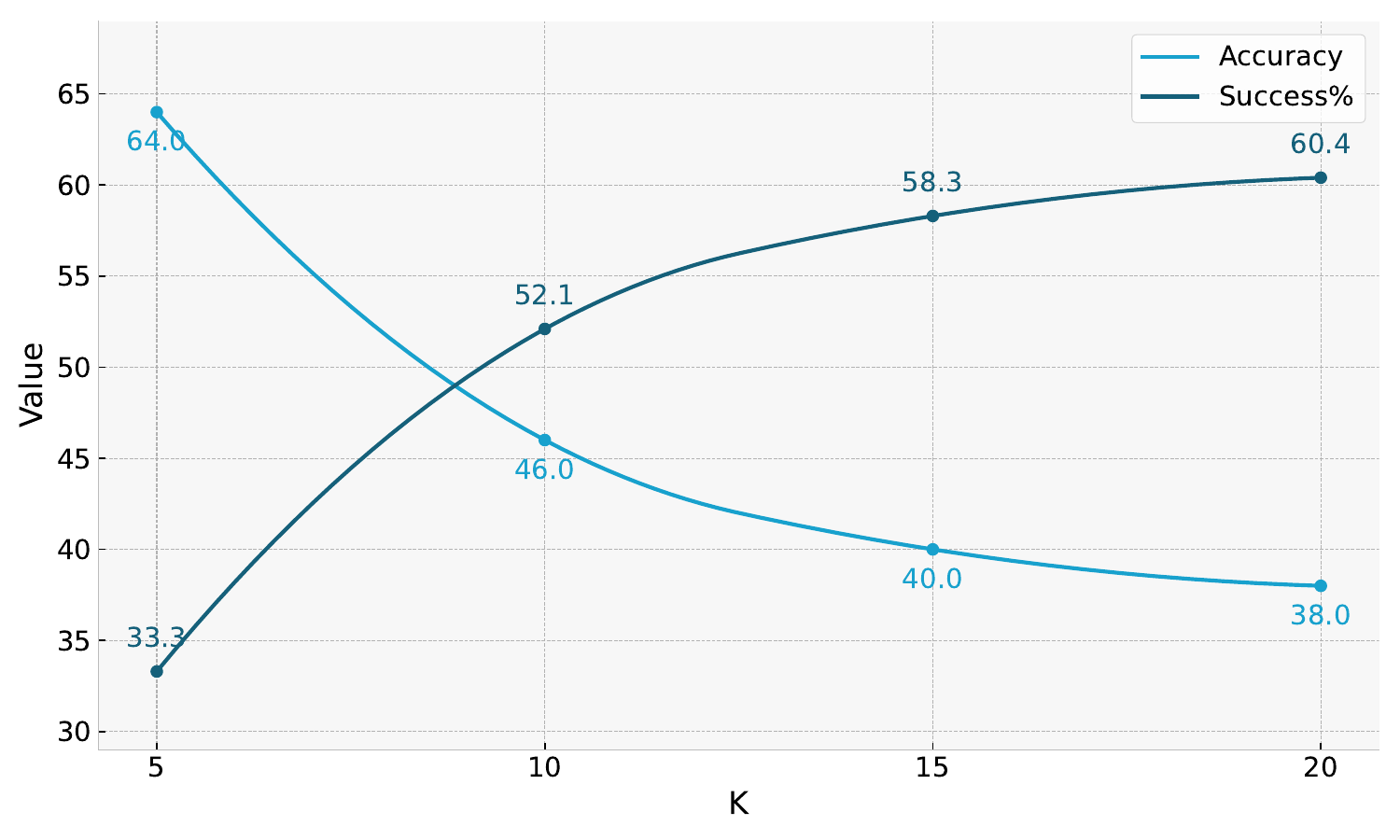}
    \caption{Accuracy and attack success rates of the adversarial examples over $50$ samples in the IMDB dataset with different $k$ values.}
    \label{fig:grid}
\end{figure}

Following the previous works \cite{Jin-etal-2020-TextFooler,Fang-etal-2023-Decision-Making}, we conducted the main experiments on the standard BERT (Bidirectional Encoder Representations from Transformers) model \cite{Devlin-etal-2019-BERT}. All victim models were pre-trained from \verb|TextAttack| \cite{Morris-etal-2020-TextAttack}, and we adopted \verb|TextAttack| to implement \textsc{LLM-Attack}’s adversarial attack process. We set the maximum modification rate $\theta$ as $0.4$, the maximum word embedding distance as $0.5$, and the USE threshold as $0.9$. Furthermore, we used ChatGLM-Pro \cite{Zeng-etal-2023-GLM} as the LLM, and set the temperature as $0$ to guarantee relatively stable outputs from the model. We set the number of synonyms obtained from the LLM as $15$ after balancing the effectiveness and efficiency of using grid search, as shown in Fig. \ref{fig:grid}. All the experiments were conducted on a single NVIDIA GeForce RTX 2080 Ti graphics card.

\subsection{Automatic Evaluation Results}

\begin{table}[t!]
    \centering
    \caption{Statistics of Datasets}
    \renewcommand\arraystretch{1.2}
    \setlength{\tabcolsep}{4mm}{\begin{tabular}{c|cccc}
        \toprule
        \textbf{Dataset} & \textbf{\#Train} & \textbf{\#Test} & \textbf{Avg. Len.} & \textbf{\#Classes} \\
        \midrule
        MR \cite{Pang-and-Lee-2005-MR} & $9$K & $1$K & $22$ & $2$ \\
        IMDB \cite{Maas-etal-2011-IMDB} & $25$K & $25$K & $280$ & $2$ \\
        Yelp \cite{Zhang-etal-2015-AG-Yelp} & $560$K & $38$K & $154$ & $2$ \\
        \bottomrule
    \end{tabular}}
    \label{tab:datasets}
\end{table}

\begin{table*}[t!]
    \centering
    \caption{Automatic Evaluation Results of LLM-Attack against Baselines}
    \renewcommand\arraystretch{1.2}
    \setlength{\tabcolsep}{5.8mm}{\begin{tabular}{l|cc|cc|cc}
        \toprule
        \textbf{Model} & \multicolumn{2}{c|}{\textbf{MR}} & \multicolumn{2}{c|}{\textbf{IMDB}} & \multicolumn{2}{c}{\textbf{Yelp}} \\
        \midrule
        BERT \cite{Devlin-etal-2019-BERT} & \multicolumn{2}{c|}{$87.8$} & \multicolumn{2}{c|}{$92.4$} & \multicolumn{2}{c}{$97.0$} \\ 
        \midrule
         & \textbf{Accuracy$\downarrow$} & \textbf{Success\%$\uparrow$} & \textbf{Accuracy$\downarrow$} & \textbf{Success\%$\uparrow$} & \textbf{Accuracy$\downarrow$} & \textbf{Success\%$\uparrow$} \\
        \midrule
        \textsc{TextFooler} \cite{Jin-etal-2020-TextFooler} & $9.0$ & $89.8$ & $1.2$ & $98.7$ & $5.0$ & $94.9$ \\
        \rowcolor{gray!15} $\quad$ \textit{w/ Constraints} & $60.2$ & $31.4$ & $22.2$ & $76.0$ & $32.8$ & $66.2$ \\
        \verb|BAE| \cite{Garg-and-Ramakrishnan-2020-BAE} & $35.8$ & $59.2$ & $30.0$ & $67.5$ & $40.0$ & $58.8$ \\
        \rowcolor{gray!15} $\quad$ \textit{w/ Constraints} & $73.6$ & $16.2$ & $58.8$ & $36.4$ & $76.2$ & $21.4$ \\
        \textsc{LLM-Attack} & $\mathbf{65.4}$ & $\mathbf{25.5}$ & $\mathbf{42.8}$ & $\mathbf{53.7}$ & $\mathbf{56.0}$ & $\mathbf{42.3}$ \\
        \bottomrule
    \end{tabular}}
    \label{tab:automatic}
\end{table*}

We first evaluated the performance of \textsc{LLM-Attack} under the automatic evaluation setting using accuracy and attack success rate. As shown in Table \ref{tab:automatic}, although there is a gap in the evaluation results of \textsc{LLM-Attack} against baselines, it is still considered as an effective attack model that achieves success rates of $25.5\%$, $53.7\%$, and $42.3\%$ on the MR, IMDB, and Yelp Review Polarity datasets, respectively. Consistent with the findings reported in the previous work \cite{Fang-etal-2023-Decision-Making}, \textsc{LLM-Attack} is potentially more effective on document-level datasets since the victim models are more likely to rely on surface clues in long documents rather than comprehending them when making predictions.

The original \textsc{TextFooler} and \verb|BAE| adversarial attack models performed effectively on the automatic evaluation metrics; however, after applying the same constraints from \textsc{LLM-Attack}, we observed dramatic performance drops in the baseline models, especially \verb|BAE|. This suggests that the original constraints are inefficient in restricting the generated adversarial examples. Due to the word embedding distance being initially taken into account while generating adversarial examples, the decrease in automatic evaluation metrics for \textsc{TextFooler} is less significant than for \verb|BAE|. \textsc{LLM-Attack} outperformed \verb|BAE| by a significant margin after applying the additional constraints, indicating that LLMs are more efficient than BERT-MLM in generating adversarial examples.

\subsection{Human and GPT-4 Evaluation Results}\label{sec:manual}

In human evaluation, we selected all the adversarial examples that successfully attack the BERT model using \textsc{TextFooler}, \verb|BAE|, and \textsc{LLM-Attack}, in which we applied all the proposed constraints. Following the criteria proposed in \cite{Dyrmishi-etal-2023-Valid-Natural}, we hired two evaluators to assess the \textit{validity} and \textit{naturalness} of the adversarial examples in five aspects: validity, suspiciousness, detectability, grammaticality, and meaning: 1) \textit{Validity} (Val.): the percentage of human judgments consistent with the ground-truth labels; 2) \textit{Suspiciousness} (Sus.): the proportion of human judgments that the samples are computer-altered; 3) \textit{Detectability} (D.)/\textit{Grammaticality} (G.)/\textit{Meaning} (M.): the percentage of the adversarial examples with the least detectability, highest grammaticality and understandability. 

Additionally, based on the previous work \cite{Liu-etal-2023-GEval} that utilised GPT-4 \cite{OpenAI-2023-GPT4} to evaluate the output of natural language generation, we also adopted GPT-4 to evaluate the detectability, grammaticality, and meaning of the adversarial examples.

Table \ref{tab:validity-sus} illustrates the human evaluation results on the validity and suspiciousness of the adversarial examples. As for the validity of the adversarial examples, we regarded the original examples as the baseline. It is evident that the adversarial examples generated by \textsc{LLM-Attack} matched better with human perception than those generated by \textsc{TextFooler} and \verb|BAE|. At the same time, the adversarial examples generated by \textsc{LLM-Attack} had the lowest likelihood of being recognised as computer-altered, which was even less than half of the baselines. In contrast, more than half of the adversarial examples generated by \textsc{TextFooler} were identified as computer-altered, indicating that those adversarial examples were unnatural and easily perceptible by humans.

\begin{figure}
    \centering
    \includegraphics[width=1\linewidth]{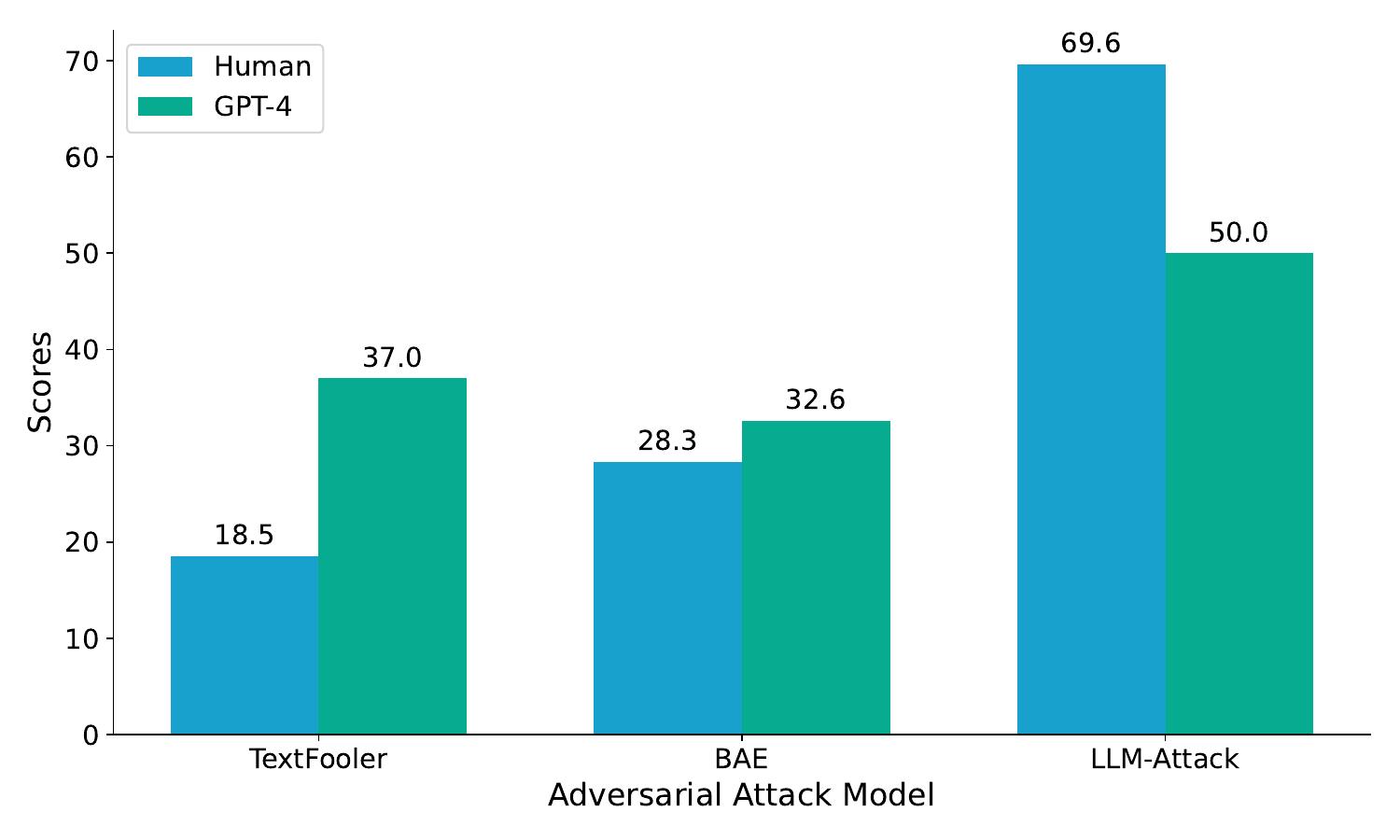}
    \caption{Percentages of the adversarial examples with the lowest detectability and the highest grammaticality and understandability selected by human evaluators and GPT-4.}
    \label{fig:dgm}
\end{figure}

We visualised the evaluation results of detectability, grammaticality, and meaning in Fig. \ref{fig:dgm}. The sum of the scores on different adversarial attack models can be greater than $1$ since some adversarial examples generated by different models are the same. \textsc{LLM-Attack} generated the most natural adversarial examples in comparison to baselines since more than half of the adversarial examples were chosen by both human evaluators and GPT-4. Consistency between humans and GPT-4 indicates the great potential of GPT-4 as an adversarial examples evaluator. \textsc{TextFooler} performed better than \verb|BAE| in validity due to the generation with word embeddings, and \verb|BAE| accomplished better in terms of naturalness because BERT-MLM pre-training makes it possible to generate  more fluent and coherent replacements. Still, there were some gaps between the validity and naturalness between the baseline models and \textsc{LLM-Attack}.

\begin{table}[t!]
    \centering
    \caption{Human Evaluation Results on Validity and Suspiciousness. Val. stands for Validity and Sus. for Suspiciousness.}
    \renewcommand\arraystretch{1.2}
    \setlength{\tabcolsep}{7.7mm}{\begin{tabular}{l|cc}
        \toprule
        \textbf{Model} & \textbf{Val.$\uparrow$} & \textbf{Sus.$\downarrow$} \\
        \midrule
        Original & $90.2\%$ & $-$ \\
        \midrule
        \textsc{TextFooler} (\textit{w/ C.}) & $78.3\%$ & $56.5\%$ \\
        \verb|BAE| (\textit{w/ C.}) & $77.2\%$ & $39.1\%$ \\
        \textsc{LLM-Attack} & $\mathbf{89.1\%}$ & $\mathbf{13.0\%}$ \\
        \bottomrule
    \end{tabular}}
    \label{tab:validity-sus}
\end{table}

\subsection{Case Study}

\begin{figure*}
    \centering
    \small
    \begin{tabular}{m{2.0cm}<{\centering}|m{12.4cm}<{}|m{.25cm}<{\centering}m{.25cm}<{\centering}m{1.1cm}<{\centering}}\toprule
        \textbf{Model} & \textbf{Text} & \textbf{V.} & \textbf{S.} & \textbf{D./G./M.} \\
        \midrule
        Original & If you can push on through the slow spots, you’ll be rewarded with some fine acting. & $-$ & $-$ & $-$ \\
        \midrule
        \textsc{TextFooler} & If you can push on through the slow spots, you’ll be {\color{PineGreen} \textbf{recompense}} with some {\color{NavyBlue} \textbf{wondrous}} acting. & $\checkmark$ & $\times$ & $\times$ \\
        \midrule
        \verb|BAE| & If you can push on through the slow spots, you’ll be rewarded with some {\color{BrickRed} {\textbf{bad}}} acting. & $\times$ & $\checkmark$ & $\times$ \\
        \midrule
        \textsc{LLM-Attack} & If you can push on through the slow spots, you’ll be {\color{NavyBlue} 
        \textbf{advantaged}} with some fine acting. & $\checkmark$ & $\checkmark$ & $\checkmark$ \\
        \bottomrule
    \end{tabular}
    \caption{Adversarial examples generated by \textsc{LLM-Attack} and baselines on the MR dataset, in which the valid word replacement, antonyms, and words that caused grammatical errors are highlighted in {\color{NavyBlue} \textbf{blue}}, {\color{BrickRed} \textbf{red}}, and {\color{PineGreen} \textbf{green}}, respectively. V. stands for Validity; S. for Suspiciousness; and D./G./M. for Detectability/Grammaticality/Meaning.}
    \label{fig:case}
\end{figure*}

Figure \ref{fig:case} displays the adversarial examples generated by \textsc{LLM-Attack} and baselines on a sample from the MR dataset. Since the attacks with \textsc{TextFooler} and \verb|BAE| failed when the same \textsc{LLM-Attack} constraints were applied, they were executed using their original implementations. Overall, the output of \textsc{LLM-Attack} was generally better than the baseline models: it substituted the word “rewarded” with its synonym “advantaged”, maintaining the semantic meaning and grammatical correctness while altering BERT’s prediction. In contrast, \textsc{TextFooler} modified the word to a ‘synonym’ with a different tense that resulted in grammatical errors, while \verb|BAE| replaced the word “fine” with its antonym “bad”. We also asked the human evaluators to verify the adversarial examples in accordance with the criteria presented in Sec. \ref{sec:manual}. Both human evaluators gave the same judgment, and they all selected the adversarial example generated by \textsc{LLM-Attack} as the optimal adversarial example, indicating that \textsc{LLM-Attack} is capable of generating more valid and natural adversarial examples.

\section{Conclusion and Future Work}

Due to the  discrete nature of human languages, generating adversarial examples that are imperceptible to human beings is a challenging task. We proposed \textsc{LLM-Attack}, which aims to generate both valid and natural adversarial examples with LLMs with high degree of undetectability. Our experimental results demonstrated the effectiveness of \textsc{LLM-Attack} against the state-of-the-art baseline models such as \textsc{TextFooler} and \verb|BAE|. It also outperformed the baselines in human and GPT-4 evaluation by a significant margin. In future, we will perform adversarial training with the generated adversarial examples and evaluate to what extent they could boost the classification performance. In addition, we will investigate the possibility of incorporating structured knowledge from quality knowledge bases in generating adversarial examples at both word and sentence levels.

\section*{Acknowledgement}

We would like to thank the human evaluators for their help in the evaluation study. This research is funded by the Postgraduate Research Scholarship (PGRS) at Xi’an Jiaotong-Liverpool University, contract number FOSA2212008, and partially supported by 2022 Jiangsu Science and Technology Programme (General Programme), contract number BK20221260.

\bibliographystyle{IEEEtran}
\bibliography{references}

\end{document}